\documentclass{article}

\PassOptionsToPackage{numbers, sort&compress}{natbib}
\usepackage[preprint]{neurips_2025}

\usepackage[utf8]{inputenc} % allow utf-8 input
\usepackage[T1]{fontenc}    % use 8-bit T1 fonts
\usepackage{hyperref}       % hyperlinks
\usepackage{url}            % simple URL typesetting
\usepackage{booktabs}       % professional-quality tables
\usepackage{amsfonts}       % blackboard math symbols
\usepackage{nicefrac}       % compact symbols for 1/2, etc.
\usepackage{microtype}      % microtypography
\usepackage{xcolor}         % colors
\usepackage{fancyhdr}       % header
\usepackage{graphicx}       % graphics
\usepackage{color}          % for comments color
\usepackage[ruled]{algorithm2e} % algorithms
\usepackage{amsmath}
\usepackage{subcaption}     % for subfigures

\usepackage{amsthm}
\theoremstyle{plain}
\newtheorem{theorem}{Theorem}

\theoremstyle{definition}

\newtheorem{constraint}[theorem]{Constraint}
\theoremstyle{remark}

\SetKwComment{Comment}{/* }{ */}

\title{FragmentRetro: A Quadratic Retrosynthetic Method Based on Fragmentation Algorithms}

\author{%
  Yu Shee \\
  Department of Chemistry\\
  Yale University\\
  New Haven, CT 06520 \\
  \texttt{yu.shee@yale.edu} \\
  \And
  Anthony M. Smaldone \\
  Department of Chemistry\\
  Yale University\\
  New Haven, CT 06520 \\
  \texttt{anthony.smaldone@yale.edu} \\
  \And
  Anton Morgunov \\
  Department of Chemistry\\
  Yale University\\
  New Haven, CT 06520 \\
  \texttt{anton@ischemist.com} \\
  \And
  Gregory W. Kyro \\
  Department of Chemistry\\
  Yale University\\
  New Haven, CT 06520 \\
  \texttt{gregory.kyro@yale.edu} \\
  \And
  Victor S. Batista \\
  Department of Chemistry\\
  Yale University\\
  New Haven, CT 06520 \\
  \texttt{victor.batista@yale.edu} \\
}

\begin{document}

\maketitle

\begin{abstract}
    Retrosynthesis, the process of deconstructing a target molecule into simpler precursors, is crucial for computer-aided synthesis planning (CASP). Widely adopted tree-search methods often suffer from exponential computational complexity. In this work, we introduce FragmentRetro, a novel retrosynthetic method that leverages fragmentation algorithms, specifically BRICS and r-BRICS, combined with stock-aware exploration and pattern fingerprint screening to achieve quadratic complexity. FragmentRetro recursively combines molecular fragments and verifies their presence in a building block set, providing sets of fragment combinations as retrosynthetic solutions. We present the first formal computational analysis of retrosynthetic methods, showing that tree search exhibits exponential complexity \(\mathcal{O}(b^h)\), DirectMultiStep scales as \(\mathcal{O}(h^6)\), and FragmentRetro achieves \(\mathcal{O}(h^2)\), where \(h\) represents the number of heavy atoms in the target molecule and \(b\) is the branching factor for tree search. Evaluations on PaRoutes, USPTO-190, and natural products demonstrate that FragmentRetro achieves high solved rates with competitive runtime, including cases where tree search fails. The method benefits from fingerprint screening, which significantly reduces substructure matching complexity. While FragmentRetro focuses on efficiently identifying fragment-based solutions rather than full reaction pathways, its computational advantages and ability to generate strategic starting candidates establish it as a powerful foundational component for scalable and automated synthesis planning. FragmentRetro code is available on \href{https://github.com/randyshee/FragmentRetro}{GitHub}
\end{abstract}

\section{Introduction}\label{sec:introduction}
\begin{figure}[htbp]
  \centering
  \includegraphics[width=\textwidth]{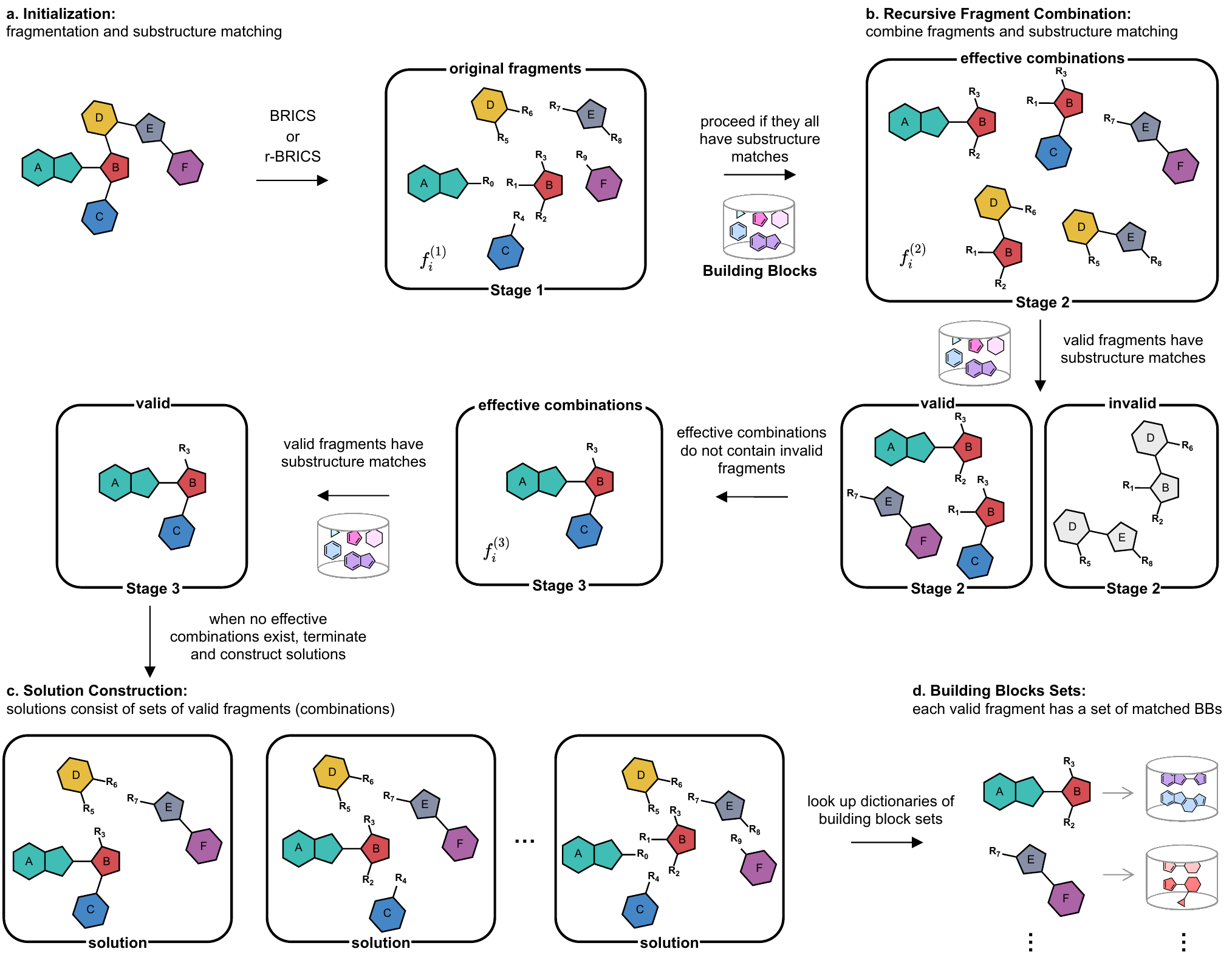}  
  \caption{The FragmentRetro algorithm. (a) Cartoon representation of an example molecule processed by BRICS or r-BRICS to yield initial fragments labeled A to F. (b) The FragmentRetro process: In Stage 1, all initial fragments have substructure matches in the stock set. In Stage 2, fragments A–B, B–C, and E–F are valid. In Stage 3, only fragment A–B–C remains valid. Fragments like A–B–D do not need to be checked, since B–D is invalid and therefore A–B–D cannot have a substructure match. There is no Stage 4, as no valid combinations of four fragments are possible in this case. (c) Possible solutions are sorted by the number of fragments, with the most efficient solution on the left. (d) Each valid fragment is associated with a subset of the stock that has substructure matches.}
  \label{fig:algorithm}
  \vspace{-1em}
\end{figure}

Retrosynthetic analysis, the deconstruction of a target molecule into simpler, typically commercially available precursors, is fundamental to chemical synthesis and drug discovery~\cite{casp_1969,grzy_2016}. Computer-aided synthesis planning (CASP) aims to automate this complex task. A conventional paradigm in CASP involves tree-search algorithms, such as Monte Carlo Tree Search (MCTS)~\cite{mcts_2018}, A* search (Retro*)~\cite{retrostar}, and Depth-First Proof Number (DFPN) search~\cite{dfpn}. These methods iteratively apply single-step retrosynthetic (SSR) models, often machine learning-based~\cite{coley_aug_2020, coley_retrosim_2017, ishida_gcn_2019,site_spec_2024}, to explore the vast combinatorial space of possible synthetic routes. While effective, a primary limitation of such tree-search approaches is their inherent exponential computational complexity with respect to tree depth, which can hinder scalability for large and complex targets. Recent advancements aim to mitigate this, for example, by incorporating graph-based neural networks~\cite{retrograph}, reinforcement learning~\cite{grasp_2022}, route context~\cite{fusionretro_2023}, or higher-level abstractions of molecular features ~\cite{higherlev_2025}. However, the underlying exponential scaling often persists in the worst-case scenario.

Alternative strategies have emerged to address these scalability challenges. DirectMultiStep (DMS) models, for instance, leverage sequence-to-sequence architectures to generate entire multi-step pathways directly, bypassing iterative search~\cite{dms_2025}. While DMS has shown promising performance, particularly on benchmarks like PaRoutes~\cite{paroutes}, and exhibits polynomial complexity, its \(\mathcal{O}(h^6)\) scaling with molecular size \(h\) can still be demanding. Other approaches, such as Double-Ended Synthesis Planning (DESP)~\cite{desp_2024}, improve efficiency for specific use cases by conducting a bidirectional tree-search when starting materials are provided by the user, though they still rely on iterative single-step predictions. This landscape motivates the exploration of fundamentally different algorithmic approaches to retrosynthesis that offer improved computational efficiency, particularly for unconstrained initial exploration, without sacrificing solution quality.

In this work, we introduce FragmentRetro, a novel retrosynthetic method that adopts a bottom-up, fragment-based strategy (Fig.~\ref{fig:algorithm}). FragmentRetro utilizes molecular fragmentation algorithms (BRICS~\cite{brics_2008} and its revision, r-BRICS~\cite{r_brics_2024}) to decompose a target molecule. It then recursively combines these fragments, systematically verifying if the combined fragments exist as substructures within a provided set of commercially available building blocks (BBs). By integrating efficient substructure matching, augmented by pattern fingerprint and property screening, FragmentRetro identifies sets of fragment combinations that reconstruct the target. A key contribution of this paper is the first formal computational complexity analysis comparing these distinct retrosynthetic paradigms. We demonstrate that while tree-search methods scale exponentially (\(\mathcal{O}(b^h)\)) and DMS polynomially (\(\mathcal{O}(h^6)\)), FragmentRetro achieves a significantly more favorable quadratic complexity (\(\mathcal{O}(h^2)\)), albeit with a linear dependency on the stock set size which is effectively managed through parallelization.

We evaluate FragmentRetro on established benchmarks, including PaRoutes and USPTO-190, as well as challenging natural products. Our results show that FragmentRetro, particularly when using r-BRICS, achieves high solved rates, competitive with or exceeding those of tree-search and DMS methods in certain scenarios, often with substantially reduced runtimes. For instance, on USPTO-190 with a large BB set, FragmentRetro with r-BRICS attains the highest solved rate (78.4\%) and demonstrates excellent parallelization for substructure matching. While FragmentRetro currently outputs sets of reconstructive fragments rather than fully elaborated reaction pathways in the form of directed acyclic graphs (DAGs), it provides a computationally efficient and scalable foundation for identifying viable precursor sets. This positions it as a powerful tool for initial synthesis exploration, potentially complementing other methods in a tiered approach to retrosynthesis. Our main contributions are: (1) the FragmentRetro algorithm, a novel fragment-based retrosynthetic method with quadratic complexity; (2) a comprehensive computational complexity analysis of major retrosynthetic paradigms; and (3) empirical validation demonstrating FragmentRetro's competitive performance and scalability.

\section{Related work}

The automation of retrosynthetic analysis has been a long-standing goal, evolving from rule-based systems to sophisticated machine learning-driven approaches. Understanding this landscape highlights the unique positioning of FragmentRetro.

\subsection{Tree-search based retrosynthesis}
The predominant approach to CASP involves constructing a search tree or graph. Early systems relied on expert-defined chemical rules~\cite{casp_1969}. Modern methods typically integrate machine learning models for SSR predictions with advanced search algorithms. These SSR models can be template-based, utilizing predefined reaction patterns~\cite{coley_aug_2020, coley_gcn_2019, coley_retrosim_2017, ishida_gcn_2019, hopfield_ssr, segler_ml_2017, localretro_2021, site_spec_2024,coley_rank_2017}, or template-free, directly predicting reactants from products~\cite{wlnet_coley_2017}. Popular search algorithms include Monte Carlo Tree Search (MCTS)~\cite{mcts_2018, aizynthfinder_2020}, A* search Retro*~\cite{retrostar}, and Proof-Number Search derivatives like DFPN~\cite{dfpn}.

Recent advancements in tree search focus on improving efficiency and prediction quality. For example, \citeauthor{retrograph} proposed RetroGraph, using graph neural networks to guide search on an explicit graph representation of intermediates to avoid redundant computations~\cite{retrograph}. Reinforcement learning techniques, such as in GRASP~\cite{grasp_2022}, train agents to learn optimal search policies. FusionRetro~\cite{fusionretro_2023} aims to improve SSR predictions by incorporating contextual information from the partially built synthetic route. \citeauthor{higherlev_2025} introduced a higher-level retrosynthesis strategy (Higherlev) that abstracts molecular details to simplify the search space, using substructure matching against BBs as a stopping criterion~\cite{higherlev_2025}. Despite these innovations, a fundamental challenge for tree-search methods is their worst-case exponential scaling with the size of the target molecule or the depth of the synthetic route, as formally analyzed in this work (Section~\ref{sec:comp_analysis}).

\subsection{Direct pathway generation and constrained search}
To circumvent the iterative nature and potential scalability issues of tree search, methods that directly generate entire multi-step synthetic pathways have been developed. DirectMultiStep (DMS)~\cite{dms_2025} employs transformer-based sequence-to-sequence models to predict a full retrosynthetic route, represented as a linearized string, from a target molecule. DMS models have demonstrated strong performance on benchmarks like PaRoutes~\cite{paroutes}, particularly in recovering known routes, and can incorporate constraints such as desired starting materials or route lengths. As shown in our complexity analysis (Section~\ref{sec:comp_analysis}), DMS exhibits polynomial complexity(\(\mathcal{O}(h^6)\)), offering an improvement over the exponential scaling of tree search.

Addressing the common real-world scenario where specific starting materials must be utilized, \citeauthor{desp_2024} introduced Double-Ended Synthesis Planning (DESP)~\cite{desp_2024}. DESP employs a bidirectional tree-search approach, simultaneously exploring retrosynthetic steps from the target and forward synthetic steps from user-specified starting materials. This method, guided by a learned synthetic distance cost function, has shown improved efficiency in solving such constrained problems compared to purely unidirectional search. However, like many tree-search methods, DESP combines explicit graph search with learned single-step predictors, and critically, it requires prior human specification of starting materials. Our work, in contrast, aims to identify potential starting material sets automatically as part of its core fragment-based exploration.

\subsection{Molecular fragmentation in chemistry}
Molecular fragmentation, the process of breaking molecules into smaller, chemically meaningful pieces, is a well-established technique in cheminformatics. It finds applications in drug design for identifying pharmacophores, in quantitative structure-activity relationship (QSAR) modeling, and as a preprocessing step for various machine learning tasks. The BRICS (Breaking of Retrosynthetically Interesting Chemical Substructures) algorithm~\cite{brics_2008} is a widely used rule-based method that cleaves molecules at bonds deemed retrosynthetically labile, effectively identifying potential synthons or key disconnections. BRICS has been noted for sometimes producing large, inflexible fragments. To address this, r-BRICS (revised BRICS) was recently developed by~\citeauthor{r_brics_2024}, extending the original rule set to enable more granular fragmentation, particularly for challenging structures like fused rings and long aliphatic chains~\cite{r_brics_2024}. Our work, FragmentRetro, leverages these fragmentation algorithms not merely for analysis, but as the core engine for a bottom-up retrosynthetic search.

\subsection{Substructure searching}
A critical component of FragmentRetro, as well as methods like Higherlev~\cite{higherlev_2025}, is substructure searching: determining if a given molecular fragment (query) exists within a larger molecule (often from a database of BBs). This is a subgraph isomorphism problem. Efficient algorithms like VF2~\cite{substruct_algo_2004} are commonly used in cheminformatics toolkits such as RDKit~\cite{rdkit_2006}. The practical performance of substructure searching can be significantly enhanced by pre-filtering candidates using molecular fingerprints (e.g., pattern fingerprints) or simple property checks (e.g., heavy atom count, ring count), as employed in FragmentRetro. While worst-case complexity for subgraph isomorphism is exponential, these heuristics and optimized algorithms make it feasible for large-scale database searching in practice~\cite{substruct_bench_2012}.

\section{Algorithm}\label{sec:algorithm}

FragmentRetro introduces a distinct paradigm for retrosynthetic analysis. Unlike top-down tree-search methods that recursively break bonds based on reaction predictions, or direct generation methods that learn entire pathway sequences, FragmentRetro operates bottom-up. It first decomposes the target into elementary fragments and then systematically explores combinations of these fragments that are present as substructures in a BB inventory. This stock-aware exploration of fragment combinations, combined with effective pruning and efficient screening, allows FragmentRetro to achieve quadratic computational complexity \(\mathcal{O}(h^2)\). While it does not directly output a sequence of reactions forming a DAG, it identifies sets of precursor fragments from which such a route could be constructed. This focus on computational efficiency and a novel search strategy distinguishes FragmentRetro from existing approaches, offering a scalable alternative for identifying potential synthetic BBs, particularly for complex targets or when exploring large chemical spaces. The formal complexity analysis presented herein further clarifies the theoretical advantages of this approach compared to established methods. In what follows, we define \(\mathcal{M}\) as the set of all molecules and \(\mathcal{B} \subset \mathcal{M}\) as the set of available BBs. For a formal statement of the retrosynthesis problem, refer to Section~\ref{sec:prob_form}.

\subsection{Fragment-based retrosynthesis}

Building upon the principles of molecular fragmentation discussed in the previous section, we introduce our fragment-based retrosynthetic approach that leverages substructure matching. This method recursively combines molecular fragments while verifying their presence in \(\mathcal{B}\). The process aims to identify retrosynthetic solutions for a given target molecule \(p^* \in \mathcal{M}\). Here, we define a retrosynthetic solution \(Q_i\) as a set of fragments that reconstruct the target molecule, i.e. \(\bigcup_{f \in Q_i} f = p^*\) where the union is taken over molecular graphs with defined attachment points (fragmented bonds).

Let \(\mathcal{F}_1 = \{ f^{(1)}_1, f^{(1)}_2, \ldots, f^{(1)}_k \}\) represent the set of \(k\) molecular fragments obtained from a fragmentation method (e.g., BRICS~\cite{brics_2008} and r-BRICS~\cite{r_brics_2024}), where each fragment \(f^{(1)}_i \in \mathcal{F}\) corresponds to a SMILES~\cite{smiles} string. These SMILES strings contain `any' atoms (dummy atoms `*', which represent any atom in SMARTS~\cite{smarts} expressions) each attaching to the fragment bond ends unless no fragmentation is performed on a given SMILES string.

Fragment combinations are defined as groups of fragments that remain bonded during fragmentation, forming a connected subgraph of neighboring fragments. For example, \(\mathcal{F}_1\) represents single fragments, while \(\mathcal{F}_n\) represents combinations of \(n\) neighboring fragments, which are still represented as single SMILES strings, possibly with dummy atoms. These combinations expand as the recursion progresses, increasing the number of fragments combined and refining the search for valid retrosynthetic solutions. FragmentRetro identifies valid retrosynthetic solutions through the following recursive process:
\begin{enumerate}
    \item \textbf{Initialization:}  For each fragment \(f^{(1)}_i \in \mathcal{F}_1\), check if  \(\exists ~ b \in \mathcal{B}\) s.t. \(f^{(1)}_i\) is a substructure of \(b\), i.e., \(f^{(1)}_i \subseteq b\). Here, strict substructure matches are done using SMARTS patterns to avoid branchings from non-fragmented sites. Also, the `any' atoms can be hydrogen atoms (no branching from the fragmented sites). We call \(f^{(1)}_i\) a valid fragment if there is a substructure match. Aromaticity and chirality must match for a candidate BB to be considered as containing the substructure of the fragment, except at the fragmented sites where stereochemistry may be undefined. If  \(\exists ~ f^{(1)}_i \in \mathcal{F}_1\) s.t. \(f^{(1)}_i \not\subseteq b\) \(\forall ~ b \in \mathcal{B}\), the process terminates immediately, as subsequent fragment combinations would also fail to have substructure matches.
    \item \textbf{Recursive Fragment Combination:} At each iteration \(n \geq 2\), obtain combinations of \(n\) neighboring fragments (connected subgraphs with \(n\) nodes) from \(\mathcal{F}_1\). Form a fragment set \(\mathcal{F}_n = \{ f^{(n)}_1, f^{(n)}_2, \ldots, f^{(n)}_{k^{(n)}}\}\), where each \(f^{(n)}_{i}\) can be written as \(f^{(1)}_{j_1} \cup \ldots \cup f^{(1)}_{j_n}\) with \(j_1, \ldots, j_n \leq k\). For each fragment \(f^{(n)}_{i} \in \mathcal{F}_n\), check if \(\exists ~ b \in \mathcal{B}\) such that \(f^{(n)}_{i} \subseteq b\). We also record the subset of \(\mathcal{B}\) that have a substructure match. When checking larger fragment combinations, we then take the intersection of these subsets, limiting the substructure check to a potentially much smaller set of BBs. One can save time by pruning fragments containing invalid fragments from previous stages to avoid redundant checks. This gives a subset of \(\mathcal{F}_n\) that we call it the effective set \(\mathcal{F}^{\prime}_n\) at stage \(n\). For example, if a combination \(f^{(1)}_{j_1} \cup f^{(1)}_{j_2}\) at stage \(n=2\) does not have substructure matches in \(\mathcal{B}\), any larger combination containing it (e.g., \(f^{(1)}_{j_1} \cup f^{(1)}_{j_2} \cup f^{(1)}_{j_3}\) at stage \(n=3\) is excluded from evaluation). This, in practice, also prevents the process of finding all connected subgraphs with \(n\) nodes, as one can find the effective combinations from the valid combinations of the \(n-1\) stage. Additionally, strategies such as fingerprint screening can be used to filter out BBs that lack the pattern fingerprints of the fragment SMILES, significantly reducing the number of BBs that need to be checked for substructure matches (it's also applicable to the substructure matching process during the \textbf{Initialization} stage). In this work, FragmentRetro not only uses fingerprint screening but also utilizes screening based on the number of heavy atoms and rings, ensuring that BBs have greater than or equal numbers of heavy atoms and rings compared to the fragment combinations (i.e. property screening).
    \item \textbf{Termination and Solution Construction:} The process terminates when either \(\mathcal{F}^{\prime}_n  = \emptyset\) (indicating no effective or valid combinations exist) or \(\mathcal{F}_n = \{p^*\}\) (or \(n=|\mathcal{F}_1|\)). The final retrosynthetic solutions \(Q = \{ Q_1, Q_2, \ldots \}\) consist of sets of fragments \(Q_i = \{ f^{(a)}_{x}, f^{(b)}_{y}, \ldots \}\), where each fragment \(f^{(a)}_{x}\) represents a valid combination of original fragments from \(\mathcal{F}_1\). Each \(Q_i\) satisfies the condition \(\bigcup_{f \in Q_i} f = p^*\). The most efficient solution minimizes the number of fragments in \(Q_i\), i.e., \(|Q_i|\).
\end{enumerate}

By iteratively expanding and pruning fragment combinations, FragmentRetro balances exhaustive exploration with computational efficiency, effectively navigating the constraints imposed by the BBs. This approach provides a robust framework for analyzing retrosynthetic pathways while deferring the selection of specific reactions to downstream analysis. The reaction selection process itself is relatively straightforward and depends heavily on the fragmentation algorithm. For instance, if the fragmentation algorithm (e.g., BRICS, as used in this work) is well-defined and consistent, it is possible to establish a mapping between specific fragment bond breaks and common reaction types associated with those bond breaks.

Unlike the explicit synthetic routes \(S\) generated in tree-search and DirectMultiStep methods, \(Q_i\) represents a set of fragments that reconstruct the target molecule \(p^*\) but does not inherently define the sequence of reaction steps. Without specifying the order of reactions, a DAG representation of the synthetic route cannot be constructed. For instance, given three fragments \(f^{(a)}_{x}\), \(f^{(b)}_{y}\), and \(f^{(c)}_{z}\), it must be determined whether \(f^{(a)}_{x}\) and \(f^{(b)}_{y}\) react first, or if all three participate in a three-component reaction. Additionally, intermediate products formed during these reactions must be explicitly defined. Once the reaction order and intermediate products are specified, the sequence of reactions \(S\) can be constructed, enabling the formation of a DAG that represents the synthetic route.

\subsection{Algorithm summary}

The FragmentRetro method employs a multi-stage, stock-aware exploration strategy for fragment combination. Each stage checks feasibility of fragment combinations, reducing the search space by pruning infeasible combinations. The algorithm is summarized in Algorithm~\ref{alg:fragment_retro} and an illustration is shown in Fig.~\ref{fig:algorithm}. Note that the effective combinations at stage \(n\) can be found by adding neighboring fragments to the valid combinations of the \(n-1\) stage. This prevents the process of finding all connected subgraphs with \(n\) nodes. 

\begin{algorithm}
\caption{FragmentRetro(\(p^*\), \(\mathcal{B}\), \(\mathcal{F}_1 = \{ f^{(1)}_1, f^{(1)}_2, \ldots, f^{(1)}_k \})\) \\
\(p^*\): target molecule, \(\mathcal{B}\): building blocks, \(\mathcal{F}_1\): initial fragment set from fragmentation algorithms}
\For{\(i \gets 1\) to \(|\mathcal{F}_1|\)}{
    \If{\(f^{(1)}_i \not\subseteq b\) for all \(b \in \mathcal{B}\)}{
        \textbf{exit} \Comment*[r]{All initial fragments should be valid to continue}
    }
}

\For{\(n \gets 2\) to \(|\mathcal{F}_1| - 1\)}{
    Obtain \(\mathcal{F}_n = \{ f^{(n)}_1, f^{(n)}_2, \ldots, f^{(n)}_{k^{(n)}}\}\) \Comment*[r]{\(n\) neighboring combinations}
    Prune to get effective combinations \(\mathcal{F}^{\prime}_n\) \Comment*[r]{Check for redundancy} 

    \If{\(\mathcal{F}^{\prime}_n = \emptyset\) or \(\mathcal{F}_n = \{p^*\}\)}{
    \Comment{No valid fragments exist or the target molecule is formed}
    \textbf{break}}

    \uIf{\(f^{(n)}_i \subseteq b\) for some \(b \in \mathcal{B}\)}{
        \Comment{Fingerprint screening on \(\mathcal{B}\) can be performed here using \(f^{(n)}_i\)} \
        Mark \(f^{(n)}_i\) as valid \Comment*[r]{Check substructure match in \(\mathcal{B}\)}
    }
    \Else{
        Mark \(f^{(n)}_i\) as invalid \Comment*[r]{Help check redundancy in the next stage}
    }
}
\Return solutions \(Q = \{ Q_1, Q_2, \ldots \}\), where \(Q_i = \{ f^{(a)}_{x}, f^{(b)}_{y}, \ldots \}\) satisfies \(\bigcup_{f \in Q_i} f = p^*\)
\label{alg:fragment_retro}
\end{algorithm}

\subsection{Computational complexity comparison}

\begin{table}[htbp]
\centering
\caption{Comparison of retrosynthetic methods.}
\label{tab:comparison}
\begin{tabular}{@{}lcccc@{}}
\toprule
\textbf{Method}        & \textbf{Complexity}  & \textbf{Feasibility Guarantee} & \textbf{Atom Mapping} & \textbf{Form DAG} \\ \midrule
% BRICS                  & \(\mathcal{O}(h)\)     & No                                  & Yes                  & No  \\
Tree Search            & \(\mathcal{O}(b^h)\)   & No                                  & Yes (template-based) & Yes \\
DirectMultiStep        & \(\mathcal{O}(h^6)\)   & No                                  & No                   & Yes \\
FragmentRetro          & \(\mathcal{O}(h^2)\)   & No                                  & Yes                  & No  \\ \bottomrule
\end{tabular}
\end{table}

The computational complexity analysis is outlined in Section~\ref{sec:comp_analysis} and summarized in Table~\ref{tab:comparison}. Here, we use the heavy atom count of a given target \(h\) as a proxy for problem difficulty. For tree-search methods, the depth of the search tree for a given target can be estimated as \(h / \Delta h\), where \(\Delta h\) is the average reduction in heavy-atom count per reaction. Therefore, tree-search methods scale exponentially as \(\mathcal{O}(b^{h / \Delta h}) = \mathcal{O}(b^h)\), where \(b\) is the branching factor of the search tree and \(\Delta h\) can be treated as constant. In comparison, DirectMultiStep models require repeated computation of attention matrices and exhibit polynomial complexity of \(\mathcal{O}(h^6)\), since the route sequence length scales as \(\mathcal{O}(h^2)\) and each attention operation scales as \(\mathcal{O}(h^4)\).

For FragmentRetro, the dominant computational cost arises from substructure matching. However, fingerprint screening eliminates the majority of non-matching candidates before the substructure check, effectively keeping this process inexpensive in practice. This introduces a constant prefactor \(C_{fp}\) representing the filtering efficiency. The cost still scales linearly with the stock set size \(|\mathcal{B}|\), and the number of possible fragment combinations scales quadratically with \(h\). Therefore, the overall complexity of FragmentRetro can be derived as \(\mathcal{O}(C_{fp} \cdot |\mathcal{B}| \cdot h^2) = \mathcal{O}(h^2)\). Note that this analysis describes the asymptotic behavior and comes with certain caveats. For example, we assume an ideal fingerprint that prevents collisions between molecular features, but in practice the fingerprint size is finite. Additionally, the worst case could occur for extremely ring-heavy target molecules, where r-BRICS fragmentation of ring bonds produces a large number of fragments and combinations, increasing the computational cost.

Table~\ref{tab:comparison} provides a comparison of FragmentRetro, tree-search, and DMS methods across key metrics, including computational complexity, feasibility guarantees, and atom mapping capabilities. Notably, none of these methods can guarantee that the predicted synthetic routes are experimentally feasible. However, each method has certain advantages that may improve the likelihood of generating viable routes. \textbf{Tree-search methods} that rely on reaction templates allow them to retrieve references and metadata from template datasets, which can help assess reaction feasibility. \textbf{DirectMultiStep} employs a data-driven approach, generating reactions that are statistically closer to experimentally validated synthetic routes. \textbf{FragmentRetro} leverages fragmentation algorithms that preferentially break bonds of retrosynthetic interest. Since these bonds are more likely to correspond to feasible reaction conditions, the resulting fragments can guide retrosynthetic planning effectively. Additionally, specific bond-breaking patterns can be mapped to known reaction types (similar to how synthons relate to actual molecules), suggesting that FragmentRetro could be extended to generate a DAG representation of synthetic routes. This remains an area for future development. Analogous to FragmentRetro's use of fragments, Higherlev also necessitates a mapping from abstracted portions to actual functional groups.

\section{Empirical validation}

\subsection{Search performance evaluation}

\begin{table}[htbp]
\centering
	\caption{Search performance on PaRoutes test sets with n\(_1\) stock or n\(_5\) stock as the stock set.}
	\begin{tabular}{lcccc}
		\toprule
		Method          & Targets         & Solved Rate & Run Time (s)\(^a\) & Clusters/Solutions\(^b\) \\
		\midrule
            MCTS\(^c\)      & Set-n\(_1\)     & 97.16\%     & 303.3           & 109 \\
                            & Set-n\(_5\)     & 96.89\%     & 365.7           & 113 \\
            Retro*\(^c\)    & Set-n\(_1\)     & 97.28\%     & 300.7           & 31  \\
                            & Set-n\(_5\)     & 97.29\%     & 349.2           & 26  \\
            DFPN\(^c\)      & Set-n\(_1\)     & 77.86\%     & 347.3           & 2   \\
                            & Set-n\(_5\)     & 67.30\%     & 297.9           & 2   \\
            \midrule
            DMS-Explorer-XL\(^{d}\)     & Set-n\(_1\)     & 80.08\%     & 14.7      & NA    \\
                                        & Set-n\(_5\)     & 79.04\%     & 16.3      & NA    \\
            \midrule
            FragmentRetro + BRICS\(^{e}\)   & Set-n\(_1\)     & 69.90\%      &  9.4 (26.2)  & 4     \\
                                            & Set-n\(_5\)     & 69.32\%      &  9.7 (27.1)  & 6     \\
            FragmentRetro + r-BRICS\(^{e}\) & Set-n\(_1\)     & 83.32\%      & 11.7 (30.2)  & 11    \\
                                            & Set-n\(_5\)     & 82.62\%      & 12.0 (32.2)  & 12    \\		
            \bottomrule
	\end{tabular}
\label{table:search_performance_paroutes}
\begin{minipage}{\textwidth}
\raggedright
\footnotesize{\(^a\)Averages over all targets.}
\\\footnotesize{\(^b\)Medians over all targets. NA indicates not available.}
\\\footnotesize{\(^c\)Data collected from the 2.0 version of PaRoutes in their GitHub repository~\cite{paroutes} under the Apache-2.0 License.}
\\\footnotesize{\(^d\)DMS models are run with a beam size of 50 on a single NVIDIA A100 GPU with half-precision floating point inference (FP16). The other methods are run with a single CPU (no parallelization)}
\\\footnotesize{\(^e\)Runtimes in parentheses are reported without property and fingerprint screening.}
\end{minipage}
\end{table}

\begin{table}[htbp]
	\caption{Search performance on USPTO-190\(^{a}\) with Buyables\(^{b}\) as the stock set}
	\centering
	\begin{tabular}{lcccc}
		\toprule
		Method\(^{c}\)                                          & Solved Rate & Run Time (s)\(^d\) \\
            \midrule
            DMS-Explorer-XL                                                 & 27.9\%      &  21.9  \\
            Original + MCTS~\cite{higherlev_2025}                           & 46.3\%      &    NA  \\
            DMS-Flash\(^e\)                                                 & 55.3\%      &  31.3  \\
            DMS-Wide\(^e\)                                                  & 56.8\%      & 105.8  \\
            Retro*-0~\cite{retrostar}\(^{f}\)                               & 73.2\%      &    NA (65.0) \\
            Higherlev + MCTS~\cite{higherlev_2025}                          & 73.7\%      &    NA  \\
            Retro*~\cite{retrostar}\(^{f}\)                                 & 76.8\%      &    NA (57.1) \\
            \midrule
            FragmentRetro + BRICS\(^{g}\)                                   & 53.2\%      & 344.0 (74.2, 49.1, 42.1)  \\
            FragmentRetro + r-BRICS\(^{g}\)                                 & 78.4\%      & 473.1 (102.1, 72.9, 64.5)  \\
            \bottomrule
	\end{tabular}
\label{table:search_performance_uspto_190}
\begin{minipage}{\textwidth}
\raggedright
\footnotesize{\(^a\) USPTO-190 is from~\cite{retrostar} (downloaded from~\cite{shee2025figshare} under the CC BY 4.0 License).}  
\\\footnotesize{\(^b\) The stock compounds (Buyables) are from~\cite{higherlev_2025} under the CC BY 4.0 License and includes 0.329M buyable building blocks from eMolecules, Sigma-Aldrich, Mcule, ChemBridge Hit2Lead, and WuXi LabNetwork.}  
\\\footnotesize{\(^c\) DMS models are run with a beam size of 50 on a single NVIDIA A100 GPU with half-precision floating point inference (FP16). The other methods are run with a single CPU (no parallelization)}  
\\\footnotesize{\(^d\) Averages over all targets. NA indicates that the runtime is not available in the original publication.} 
\\\footnotesize{\(^e\) Uses step counts from 2 to 8 (total of 7 DMS model runs per compound).}
\\\footnotesize{\(^f\) Evaluated in this work using Retro*'s~\cite{retrostar} official repository. The original checkpoint is used with a maximum of 500 iterations. Runtimes in parentheses are first-solution times, as the repository lacks search continuation.}
\\\footnotesize{\(^g\) Runtimes in parentheses are from parallelization with 5, 10, and 20 CPU cores (during substructure matching).}
\end{minipage}
\vspace{-1em}
\end{table}

Table~\ref{table:search_performance_paroutes} presents the search performance of various methods on the PaRoutes test sets n\(_1\) and n\(_5\). All methods were evaluated using a single CPU/core (no parallelization), except for the DMS models, which were run on an NVIDIA A100 GPU. The stock set for each test follows the SMs defined in n\(_1\) or n\(_5\), as done in~\cite{paroutes}. Among the evaluated methods, Retro*, a tree-search approach, achieved the highest solved rate. The DMS Explorer-XL model~\cite{dms_2025} exhibited competitive performance, solving a significant portion of the test set while providing a 20x speedup compared to tree-search methods. Similarly, FragmentRetro demonstrated decent performance, particularly with r-BRICS fragmentation, which outperformed DMS in solved rate while achieving a 30x speedup over tree-search methods. For FragmentRetro, a target is considered solved if at least one solution is found, where the solution corresponds to a set of valid fragments that can reconstruct the target compound, since the method does not output DAGs. Table~\ref{table:search_performance_paroutes} also reports the median number of solutions. Since the PaRoutes paper~\cite{paroutes} clustered synthetic routes using the method in~\cite{route_dist_2022} to estimate route diversity, we compare the number of solutions from FragmentRetro to the number of clusters from PaRoutes. However, these metrics are not directly comparable, as a single FragmentRetro solution can correspond to multiple synthetic routes—each valid fragment combination may have substructure matches with different BBs.  

Table~\ref{table:search_performance_uspto_190} summarizes the search performance of different methods on the USPTO-190 dataset, using the Buyables stock set (0.329M BBs) provided by the Higherlev study~\cite{higherlev_2025}. This stock set is a more practical choice than the eMolecule screening compounds (23.1M compounds) used in Retro*~\cite{retrostar}, as many screening compounds require custom synthesis. It is important to note that 13 compounds from USPTO-190 appear as target compounds in the single-step training dataset for Higherlev, while 47 compounds appear as targets or intermediates in the PaRoutes training set used by the DMS models. The DMS study reports solved rates both with and without these 47 compounds, showing minor differences. We expect similar results for Higherlev. FragmentRetro with BRICS fragmentation achieved a comparable solved rate to the best-performing DMS models but did not surpass Higherlev. However, with r-BRICS fragmentation, FragmentRetro achieved the highest solved rate among all methods. Despite this, the runtime of FragmentRetro is similar to that of tree-search methods in Table~\ref{table:search_performance_paroutes}, as execution time scales with the stock set size (\(|\mathcal{B}|\)), even with the prefactor \(C_{fp}\). This is due to substructure matching being the dominant computational bottleneck, as each candidate BB must be checked against the fragments. Fortunately, these checks are independent and can be efficiently parallelized across multiple CPU cores. To demonstrate this scalability, we also report FragmentRetro runtime using 5, 10, and 20 CPU cores. The results show that parallelization significantly reduces runtime, achieving speeds comparable to or better than DMS models. Notably, 5 and 10 CPU cores exhibit near-perfect parallelization efficiency, with speedups of 4.6x and 6.9x, respectively. However, with 20 CPU cores, the speedup is only 8.2x, indicating diminishing returns. This suggests that at this stock set size, the tradeoff between parallelization efficiency and overhead is optimal around 10 CPU cores.

\subsection{Case studies}

\begin{figure}[htbp]
  \centering
  \includegraphics[width=0.9\textwidth]{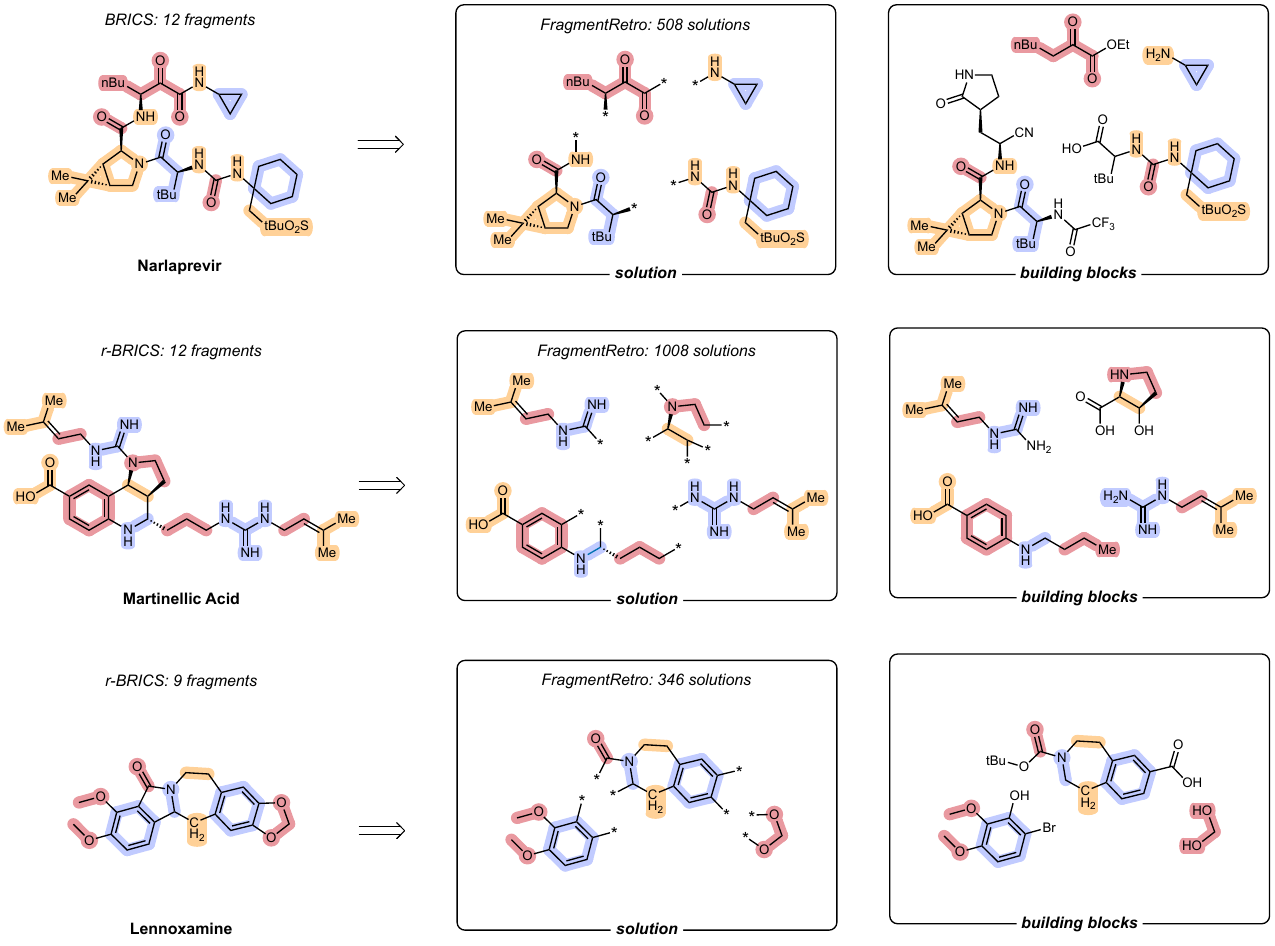}  
  \caption{FragmentRetro evaluation on Narlaprevir, Martinellic Acid, and Lennoxamine. The fragments from BRICS and r-BRICS are highlighted. One solution for each compound from FragmentRetro is shown. The highlighted fragments remain highlighted in both the solutions and corresponding building blocks, even when some BRICS or r-BRICS cleavage sites are not fragmented in the solutions.}
  \label{fig:natural_products}
  \vspace{-1em}
\end{figure}

To demonstrate the practical utility of FragmentRetro, we evaluate drugs and natural products, as shown in Fig.~\ref{fig:natural_products} (and Section~\ref{sec:add_case}). These targets are selected from the Higherlev study~\cite{higherlev_2025} and represent different levels of retrosynthetic difficulty: Narlaprevir, which can be solved without a higher-level strategy; Martinellic Acid, which requires a higher-level strategy to be solved; and Lennoxamine, which is not solvable by either approach. For Narlaprevir, we apply BRICS fragmentation rules, while for Martinellic Acid and Lennoxamine, we use r-BRICS since these compounds contain fused rings that BRICS would otherwise leave intact. FragmentRetro generates hundreds of solutions for each of these compounds. Fig.~\ref{fig:natural_products} presents a representative solution for each compound, selected to minimize the number of fragment combinations.  

Each fragment combination can match multiple BBs; Fig.~\ref{fig:natural_products} displays one representative building block per fragment combination. Some BBs are directly compatible for coupling, while others require additional preparation steps, such as functional group interconversion or protective group manipulation. As a result, the number of fragments in a solution does not directly correlate with the number of synthesis steps, similar to how higher-level abstraction methods operate.  

\section{Conclusion}  

FragmentRetro provides a scalable framework for fragment-based retrosynthesis, achieving quadratic complexity through stock-aware fragment exploration and pruning. While FragmentRetro does not explicitly return a DAG, it performs the same core task as other retrosynthetic methods: identifying chemically valid decompositions of a target molecule into known precursors. Its outputs provide sets of fragment combinations that capture aspects of both single-step (individual disconnections) and multistep (multiple simultaneous combinations) approaches, although the method does not define the exact sequence of reactions or BBs. DAG construction is a post-processing step that depends on the fragment connectivity and can be layered atop our approach. Therefore, comparing the search complexity of FragmentRetro to tree search and DMS methods is both appropriate and informative. Unlike DMS and tree-search methods that do not rely on template-based single-step models, FragmentRetro remains compatible with atom mapping tools, making it a flexible approach for retrosynthetic planning. However, FragmentRetro is not a route-level synthesis planning tool in its current form. Mapping solutions and BBs to fully elaborated synthetic routes (forming a DAG) would be required to enable direct route quality comparisons across different methods, potentially using top-\(k\) accuracy as an evaluation metric. Future work may focus on integrating cost-aware metrics and experimental validation to further refine its applicability.

This work also presents the first formal computational analysis of different retrosynthetic approaches. We establish that tree-search methods scale exponentially as \(\mathcal{O}(b^{h})\), DMS has a polynomial complexity of \(\mathcal{O}(h^{6})\), and FragmentRetro achieves \(\mathcal{O}(h^{2})\) complexity, with the caveat that its runtime scales linearly with the stock set size. However, we show that substructure matching exhibits strong parallelization efficiency, particularly when applied to large building block datasets such as Buyables. Given the trade-offs between computational cost and solved rates, an efficient retrosynthesis pipeline could adopt a tiered approach: first applying FragmentRetro for rapid exploration, then leveraging DMS for higher route quality, and finally resorting to tree-search methods when higher solved rates are required. Overall, FragmentRetro represents a promising direction for fragment-based retrosynthesis, offering a computationally efficient option that complements existing methods.

\section*{Code and Data Availability}
The data, code, and accompanying documentation for this work are available at \href{https://github.com/randyshee/FragmentRetro}{https://github.com/randyshee/FragmentRetro}

\section*{Acknowledgments}
The authors acknowledge a generous allocation of high-performance computing time from NERSC. The development of the methodology was supported by the NSF CCI grant (VSB, Award Number 2124511).

% \clearpage
\bibliography{fragmentrefs} 

@article{coley_rank_2017,
author = {Coley, Connor W. and Barzilay, Regina and Jaakkola, Tommi S. and Green, William H. and Jensen, Klavs F.},
title = {Prediction of Organic Reaction Outcomes Using Machine Learning},
journal = {ACS Central Science},
volume = {3},
number = {5},
pages = {434-443},
year = {2017},
doi = {10.1021/acscentsci.7b00064},
note ={PMID: 28573205},
URL = {https://doi.org/10.1021/acscentsci.7b00064},
eprint = {https://doi.org/10.1021/acscentsci.7b00064}
}

@inproceedings{wlnet_coley_2017,
  author       = {Wengong Jin and Connor W. Coley and Regina Barzilay and Tommi S. Jaakkola},
  editor       = {Isabelle Guyon and Ulrike von Luxburg and Samy Bengio and Hanna M. Wallach and Rob Fergus and S. V. N. Vishwanathan and Roman Garnett},
  title        = {Predicting Organic Reaction Outcomes with Weisfeiler-Lehman Network},
  booktitle    = {Advances in Neural Information Processing Systems 30: Annual Conference on Neural Information Processing Systems 2017, December 4-9, 2017, Long Beach, CA, {USA}},
  pages        = {2607--2616},
  year         = {2017},
  url          = {https://proceedings.neurips.cc/paper/2017/hash/ced556cd9f9c0c8315cfbe0744a3baf0-Abstract.html},
  bibsource    = {dblp computer science bibliography, https://dblp.org}
}

@article{coley_retrosim_2017,
author = {Coley, Connor W. and Rogers, Luke and Green, William H. and Jensen, Klavs F.},
title = {Computer-Assisted Retrosynthesis Based on Molecular Similarity},
journal = {ACS Central Science},
volume = {3},
number = {12},
pages = {1237-1245},
year = {2017},
doi = {10.1021/acscentsci.7b00355},
note ={PMID: 29296663},
URL = {https://doi.org/10.1021/acscentsci.7b00355},
eprint = {https://doi.org/10.1021/acscentsci.7b00355}
}

@article{coley_gcn_2019,
author ="Coley, Connor W. and Jin, Wengong and Rogers, Luke and Jamison, Timothy F. and Jaakkola, Tommi S. and Green, William H. and Barzilay, Regina and Jensen, Klavs F.",
title  ="A graph-convolutional neural network model for the prediction of chemical reactivity",
journal  ="Chem. Sci.",
year  ="2019",
volume  ="10",
issue  ="2",
pages  ="370-377",
publisher  ="The Royal Society of Chemistry",
doi  ="10.1039/C8SC04228D",
url  ="http://dx.doi.org/10.1039/C8SC04228D"}

@article{coley_aug_2020,
author = {Fortunato, Michael E. and Coley, Connor W. and Barnes, Brian C. and Jensen, Klavs F.},
title = {Data Augmentation and Pretraining for Template-Based Retrosynthetic Prediction in Computer-Aided Synthesis Planning},
journal = {Journal of Chemical Information and Modeling},
volume = {60},
number = {7},
pages = {3398-3407},
year = {2020},
doi = {10.1021/acs.jcim.0c00403},
note ={PMID: 32568548},
URL = {https://doi.org/10.1021/acs.jcim.0c00403},
eprint = {https://doi.org/10.1021/acs.jcim.0c00403}
}

@article{mcts_2018, 
title={Planning Chemical Syntheses with deep neural networks and symbolic AI},
volume={555}, 
DOI={10.1038/nature25978}, 
number={7698}, 
journal={Nature}, 
author={Segler, Marwin H. and Preuss, Mike and Waller, Mark P.}, 
year={2018}, 
month={Mar}, 
pages={604–610}}

@article{segler_ml_2017,
author = {Segler, Marwin H. S. and Waller, Mark P.},
title = {Neural-Symbolic Machine Learning for Retrosynthesis and Reaction Prediction},
journal = {Chemistry – A European Journal},
volume = {23},
number = {25},
pages = {5966-5971},
doi = {https://doi.org/10.1002/chem.201605499},
url = {https://chemistry-europe.onlinelibrary.wiley.com/doi/abs/10.1002/chem.201605499},
eprint = {https://chemistry-europe.onlinelibrary.wiley.com/doi/pdf/10.1002/chem.201605499},
year = {2017}
}

@article{grzy_2016,
author = {Szymkuć, Sara and Gajewska, Ewa P. and Klucznik, Tomasz and Molga, Karol and Dittwald, Piotr and Startek, Michał and Bajczyk, Michał and Grzybowski, Bartosz A.},
title = {Computer-Assisted Synthetic Planning: The End of the Beginning},
journal = {Angewandte Chemie International Edition},
volume = {55},
number = {20},
pages = {5904-5937},
doi = {https://doi.org/10.1002/anie.201506101},
url = {https://onlinelibrary.wiley.com/doi/abs/10.1002/anie.201506101},
eprint = {https://onlinelibrary.wiley.com/doi/pdf/10.1002/anie.201506101},
year = {2016}
}

@inproceedings{retrostar,
  title={Retro*: Learning Retrosynthetic Planning with Neural Guided A* Search},
  author={Chen, Binghong and Li, Chengtao and Dai, Hanjun and Song, Le},
  booktitle={The 37th International Conference on Machine Learning (ICML 2020)},
  year={2020}
}

@inproceedings{dfpn,
 author = {Kishimoto, Akihiro and Buesser, Beat and Chen, Bei and Botea, Adi},
 booktitle = {Advances in Neural Information Processing Systems},
 editor = {H. Wallach and H. Larochelle and A. Beygelzimer and F. d\textquotesingle Alch\'{e}-Buc and E. Fox and R. Garnett},
 pages = {},
 publisher = {Curran Associates, Inc.},
 title = {Depth-First Proof-Number Search with Heuristic Edge Cost and Application to Chemical Synthesis Planning},
 url = {https://proceedings.neurips.cc/paper_files/paper/2019/file/4fc28b7093b135c21c7183ac07e928a6-Paper.pdf},
 volume = {32},
 year = {2019}
}

@article{hopfield_ssr,
author = {Seidl, Philipp and Renz, Philipp and Dyubankova, Natalia and Neves, Paulo and Verhoeven, Jonas and Wegner, J{\"o}rg K. and Segler, Marwin and Hochreiter, Sepp and Klambauer, G{\"u}nter},
title = {Improving Few- and Zero-Shot Reaction Template Prediction Using Modern Hopfield Networks},
journal = {Journal of Chemical Information and Modeling},
volume = {62},
number = {9},
pages = {2111-2120},
year = {2022},
doi = {10.1021/acs.jcim.1c01065},
note ={PMID: 35034452},
URL = {https://doi.org/10.1021/acs.jcim.1c01065},
eprint = {https://doi.org/10.1021/acs.jcim.1c01065}
}

@inproceedings{retrograph,
author = {Xie, Shufang and Yan, Rui and Han, Peng and Xia, Yingce and Wu, Lijun and Guo, Chenjuan and Yang, Bin and Qin, Tao},
title = {RetroGraph: Retrosynthetic Planning with Graph Search},
year = {2022},
isbn = {9781450393850},
publisher = {Association for Computing Machinery},
address = {New York, NY, USA},
url = {https://doi.org/10.1145/3534678.3539446},
doi = {10.1145/3534678.3539446},
booktitle = {Proceedings of the 28th ACM SIGKDD Conference on Knowledge Discovery and Data Mining},
pages = {2120–2129},
numpages = {10},
location = {Washington DC, USA},
series = {KDD '22}
}

@inproceedings{grasp_2022,
 author = {Yu, Yemin and Wei, Ying and Kuang, Kun and Huang, Zhengxing and Yao, Huaxiu and Wu, Fei},
 booktitle = {Advances in Neural Information Processing Systems},
 editor = {S. Koyejo and S. Mohamed and A. Agarwal and D. Belgrave and K. Cho and A. Oh},
 pages = {10257--10268},
 publisher = {Curran Associates, Inc.},
 title = {GRASP: Navigating Retrosynthetic Planning with Goal-driven Policy},
 url = {https://proceedings.neurips.cc/paper_files/paper/2022/file/42beaab8aa8da1c77581609a61eced93-Paper-Conference.pdf},
 volume = {35},
 year = {2022}
}

@misc{desp_2024,
      title={Double-Ended Synthesis Planning with Goal-Constrained Bidirectional Search}, 
      author={Kevin Yu and Jihye Roh and Ziang Li and Wenhao Gao and Runzhong Wang and Connor W. Coley},
      year={2024},
      eprint={2407.06334},
      archivePrefix={arXiv},
      primaryClass={cs.AI},
      url={https://arxiv.org/abs/2407.06334}, 
}

@Article{paroutes,
author ="Genheden, Samuel and Bjerrum, Esben",
title  ="PaRoutes: towards a framework for benchmarking retrosynthesis route predictions",
journal  ="Digital Discovery",
year  ="2022",
volume  ="1",
issue  ="4",
pages  ="527-539",
publisher  ="RSC",
doi  ="10.1039/D2DD00015F",
url  ="http://dx.doi.org/10.1039/D2DD00015F"}

@article{casp_1969,
	title = {Computer-{Assisted} {Design} of {Complex} {Organic} {Syntheses}: {Pathways} for molecular synthesis can be devised with a computer and equipment for graphical communication.},
	volume = {166},
	issn = {0036-8075, 1095-9203},
	url = {https://www.science.org/doi/10.1126/science.166.3902.178},
	doi = {10.1126/science.166.3902.178},
	number = {3902},
	journal = {Science},
	author = {Corey, E. J. and Wipke, W. Todd},
	year = {1969},
	pages = {178--192}
}

@article{ishida_gcn_2019,
  title={Prediction and interpretable visualization of retrosynthetic reactions using graph convolutional networks},
  author={Ishida, Shoichi and Terayama, Kei and Kojima, Ryosuke and Takasu, Kiyosei and Okuno, Yasushi},
  journal={Journal of chemical information and modeling},
  volume={59},
  number={12},
  pages={5026--5033},
  year={2019},
  publisher={ACS Publications}
}

@article{localretro_2021,
  title={Deep retrosynthetic reaction prediction using local reactivity and global attention},
  author={Chen, Shuan and Jung, Yousung},
  journal={JACS Au},
  volume={1},
  number={10},
  pages={1612--1620},
  year={2021},
  publisher={ACS Publications}

}

@article{site_spec_2024,
  title={Site-specific template generative approach for retrosynthetic planning},
  author={Shee, Yu and Li, Haote and Zhang, Pengpeng and Nikolic, Andrea M and Lu, Wenxin and Kelly, H Ray and Manee, Vidhyadhar and Sreekumar, Sanil and Buono, Frederic G and Song, Jinhua J and others},
  journal={Nature Communications},
  volume={15},
  number={1},
  pages={7818},
  year={2024},
  publisher={Nature Publishing Group UK London}
}

@article{higherlev_2025,
  title={Higher-level Strategies for Computer-Aided Retrosynthesis},
  author={Roh, Jihye and Joung, Joonyoung F and Yu, Kevin and Tu, Zhengkai and Bartholomew, G Logan and Santiago-Reyes, Omar A and Fong, Mun Hong and Sarpong, Richmond and Reisman, Sarah E and Coley, Connor W},
  year={2025},
  url={https://chemrxiv.org/engage/chemrxiv/article-details/67a367196dde43c908af44a1},
  urldate={2025-03-04},
  journal={ChemRxiv preprint},
}

@inproceedings{fusionretro_2023,
  title={FusionRetro: molecule representation fusion via in-context learning for retrosynthetic planning},
  author={Liu, Songtao and Tu, Zhengkai and Xu, Minkai and Zhang, Zuobai and Lin, Lu and Ying, Rex and Tang, Jian and Zhao, Peilin and Wu, Dinghao},
  booktitle={International Conference on Machine Learning},
  pages={22028--22041},
  year={2023},
  organization={PMLR}
}

@article{brics_2008,
  title={On the art of compiling and using'drug-like'chemical fragment spaces},
  author={Degen, Jorg and Wegscheid-Gerlach, Christof and Zaliani, Andrea and Rarey, Matthias},
  journal={ChemMedChem},
  volume={3},
  number={10},
  pages={1503},
  year={2008}
}

@article{substruct_bench_2012,
  title={Systematic benchmark of substructure search in molecular graphs-From Ullmann to VF2},
  author={Ehrlich, Hans-Christian and Rarey, Matthias},
  journal={Journal of cheminformatics},
  volume={4},
  pages={1--17},
  year={2012},
  publisher={Springer}
}

@misc{rdkit_2006,
  title={RDKit: Open-source cheminformatics},
  author={Landrum, Greg and others},
  year={2006},
  publisher={Zenodo}
}

@article{substruct_algo_2004,
  title={A (sub) graph isomorphism algorithm for matching large graphs},
  author={Cordella, Luigi P and Foggia, Pasquale and Sansone, Carlo and Vento, Mario},
  journal={IEEE transactions on pattern analysis and machine intelligence},
  volume={26},
  number={10},
  pages={1367--1372},
  year={2004},
  publisher={IEEE}
}

@article{r_brics_2024,
  title={r-BRICS--A Revised BRICS Module That Breaks Ring Structures and Carbon Chains},
  author={Zhang, Leili and Rao, Vasumitra and Cornell, Wendy},
  journal={ChemMedChem},
  volume={19},
  number={4},
  pages={e202300202},
  year={2024},
  publisher={Wiley Online Library}
}

@article{dms_2025,
author = {Shee, Yu and Morgunov, Anton and Li, Haote and Batista, Victor S.},
title = {DirectMultiStep: Direct Route Generation for Multistep Retrosynthesis},
journal = {Journal of Chemical Information and Modeling},
volume = {0},
number = {0},
pages = {null},
year = {0},
doi = {10.1021/acs.jcim.4c01982},
note ={PMID: 40197023},
URL = {https://doi.org/10.1021/acs.jcim.4c01982},
eprint = {https://doi.org/10.1021/acs.jcim.4c01982}
}

@misc{shee2025figshare,
    author = {Yu Shee and Anton Morgunov},
    title = {Data for ``{DirectMultiStep}: {Direct} {Route} {Generation} for {Multistep} {Retrosynthesis}''},
    howpublished = {\url{https://figshare.com/articles/dataset/Data_for_DirectMultiStep_Direct_Route_Generation_for_Multistep_Retrosynthesis_/28629470}},
year={2025},
    doi = {"10.6084/m9.figshare.28629470.v1"},
    note = {Accessed: 2025-03-25}
}

@misc{smiles,
  title = {Daylight Theory: {SMILES}},
  howpublished = {\url{https://www.daylight.com/dayhtml/doc/theory/theory.smiles.html}},
  note = {Accessed: 2025-03-10}
}

@misc{smarts,
  title = {Daylight Theory: {SMARTS}},
  howpublished = {\url{https://www.daylight.com/dayhtml/doc/theory/theory.smarts.html}},
  note = {Accessed: 2025-03-10}
}

@article{route_dist_2022,
  title={Fast prediction of distances between synthetic routes with deep learning},
  author={Genheden, Samuel and Engkvist, Ola and Bjerrum, Esben},
  journal={Machine Learning: Science and Technology},
  volume={3},
  number={1},
  pages={015018},
  year={2022},
  publisher={IOP Publishing}
}

@article{aizynthfinder_2020,
  title={AiZynthFinder: a fast, robust and flexible open-source software for retrosynthetic planning},
  author={Genheden, Samuel and Thakkar, Amol and Chadimov{\'a}, Veronika and Reymond, Jean-Louis and Engkvist, Ola and Bjerrum, Esben},
  journal={Journal of cheminformatics},
  volume={12},
  number={1},
  pages={70},
  year={2020},
  publisher={Springer}
}
%%%%%%%%%%%%%%%%%%%%%%%%%%%%%%%%%%%%%%%%%%%%%%%%%%%%%%%%%%%%
\clearpage

\appendix

\section{Problem formulation}\label{sec:prob_form}
In this work, we consider two existing retrosynthetic methods—template-based tree search and DMS—as well as our proposed FragmentRetro method. While their specific approaches differ, their general objective remains the same: to generate synthetic routes that reconstruct a target molecule from available BBs.

\subsection{Tree search}
Here, we adopt the notation introduced in the DESP paper~\cite{desp_2024} and begin by focusing on template-based tree-search retrosynthetic approaches. These approaches are compatible with any single-step retrosynthesis machine learning model, although certain models may not provide atom mappings for reactions. Let \(\mathcal{M}\) denote the set of all molecules, \(\mathcal{R}\) the set of all reactions, and \(\mathcal{T}\) the set of all reaction templates. A \textit{reaction} \(R_i \in \mathcal{R}\) is represented as a tuple \((r_i, p_i, t_i)\), where \(r_i \subset \mathcal{M}\) is the set of reactants, \(p_i \in \mathcal{M}\) is the product, and \(t_i \in \mathcal{T}\) is the corresponding retro template. A \textit{retro template} \(t\) is a function \(t: \mathcal{M} \rightarrow 2^\mathcal{M}\) that maps a product molecule \(p_i\) to a set of potential reactants \(r_i\), such that for any given reaction \(R_i\), \(r_i \in t_i(p_i)\).

Given a target molecule \(p^* \in \mathcal{M}\) and a set of BBs \(\mathcal{B} \subset \mathcal{M}\), the goal of synthesis planning is to identify a \textit{valid synthetic route}, defined as a collection of reactions \(S = \{ R_1, \ldots, R_n \}\) that fulfills the following constraints:

\begin{constraint}[Synthesize all non-BBs]
    % \(\forall ~ i: m \in r_i, m \notin \mathcal{B} \implies \exists ~ j \text{ s.t.   } m = p_j\);
    % \(\forall R_i \in S : \forall m \in r_i,\; m \notin \mathcal{B} \Rightarrow \exists R_j \in S \text{ s.t. } m = p_j\)
    \(\forall R_i \in S,\; \forall m \in r_i,\; m \notin \mathcal{B} \Rightarrow \exists R_j \in S \text{ s.t. } m = p_j\)
    \label{constr:1}
\end{constraint}
\begin{constraint}[Target is final molecule synthesized]
    % \(\exists ~ j \text{ s.t. } p_j = p^*, \forall ~ i: p^* \notin r_i\);
    % \(\exists R_j \in S \text{ s.t. } (p_j = p^* \land (\forall R_k \in S \Rightarrow p^* \notin r_k))\)
    \(\exists R_j \in S \text{ s.t. } p_j = p^* \land \forall R_k \in S,\; p^* \notin r_k\)
    \label{constr:2}
\end{constraint}

The reactions in \(S\) form a directed acyclic graph (DAG), where each product \(p_i\) is mapped to a node. This node has a directed edge to a node representing reaction \(R_i\), which in turn has directed edges to nodes representing the reactants \(r_i\). The DAG structure ensures a logical order of reactions and prevents cyclic dependencies.

For tree-search methods that do not rely on template-based single-step predictions or do not provide atom mappings, a reaction \(R_i\) is represented as a tuple \((r_i, p_i)\), where \(r_i \in \theta(p_i)\). Here, \(\theta\) denotes the neural network predicting the set of possible reactants for a given product \(p_i\). One could argue that reactions defined with templates (or atom mappings) are \textit{valid} reactions, while those without templates may not be valid. However, even the so-called valid reactions are not guaranteed to work experimentally. Therefore, the actual retrosynthesis problem is more complex than the search problem formulated here.

\subsection{DirectMultiStep}\label{subsection:problem_DirectMultiStep}
DMS methods~\cite{dms_2025} approach the retrosynthesis problem differently from tree-search methods. Instead of representing synthetic routes as sets of reactions, DMS represents an entire synthetic route as a single string in a recursive dictionary format. ML models are trained to output these route strings given a target molecule \(p^*\) as input. Valid routes that follow the same two constraints can similarly be represented as a set of reactions \(S = \{ R_1, \ldots, R_n \}\) that forms a DAG. The \(R_i\) here is a tuple \((r_i, p_i)\) without a template.

A route \(s\) can also be programmatically represented using a recursive dictionary format:
\begin{verbatim}
class DMSDict(TypedDict, total=False):
    smiles: str
    children: List["DMSDict"]
\end{verbatim}

Here, the field \texttt{smiles} contains the SMILES representation of a molecule, and the field \texttt{children} contains the subroutes leading to the reactants for this molecule. For example, a two-step synthetic route for a molecule might have the following structure:
\begin{verbatim}
{`smiles': `target_molecule',
 `children': [{`smiles': `reactant_1'},
              {`smiles': `reactant_2',
               `children': [{`smiles': `reactant_3'}]}]}
\end{verbatim}

The neural network in DMS is trained to generate such recursive representations, predicting both the products and reactants for each step of the synthesis as a next token prediction task. This eliminates the need for explicit tree traversal and instead focuses on learning the structure of valid experimental routes~\cite{paroutes} directly.

\section{Computational complexity analysis}\label{sec:comp_analysis}

This section provides the computational complexity analysis for the three retrosynthetic planning methods: tree search, DMS, and FragmentRetro. For all methods, we denote the number of heavy atoms in the target molecule as \(h\) and the branching factor of the search tree as \(b\). Additional parameters are defined as needed.

\subsection{Tree search complexity analysis}\label{subsection:complexity_tree_search}
Assuming the computational complexity of single-step retrosynthetic predictions is constant with respect to \(h\), the tree-search process is the main consideration in the complexity analysis here. The branching factor \(b\) corresponds to the number of possible reactions or templates at each level. The depth of the tree is approximately \(h / \Delta h\), where \(\Delta h\) is the average reduction in heavy atom count per reaction. Consequently, the tree search explores \(b^{h / \Delta h}\) nodes. 

Considering that other operations, such as verifying molecule availability in a commercial compound stock set, are also constant-time operations, the overall complexity of tree search is dominated by the exponential factor: \(\mathcal{O}(b^{h / \Delta h}) = \mathcal{O}(b^h)\) as \(\Delta h\) is constant with respect to \(h\). This also applies to tree-search methods that use best-first search, as the search space for best-first search remains exponential.

\subsection{DirectMultiStep complexity analysis}\label{subsection:complexity_dms}
DMS employs an attention-based model (a transformer) to predict a synthesis route given a target compound. The attention mechanism operates with a complexity of \(\mathcal{O}(l^2 d)\), where \(l\) is the sequence length and \(d\) is the embedding dimension. In this context, sequences can represent either molecular SMILES strings or synthesis routes, which are represented as strings of recursive dictionaries (see Section~\ref{subsection:problem_DirectMultiStep}).  

The sequence length of a SMILES string scales linearly with the number of heavy atoms \(h\). Since a synthesis route comprises the target SMILES string along with the SMILES representations of its precursors (including intermediates and SMs), its sequence length is given by \(\mathcal{O}(h \cdot h / \Delta h) = \mathcal{O}(h^2)\), where \(h / \Delta h\) corresponds to the number of reaction steps (analogous to tree depth in Section~\ref{subsection:complexity_tree_search}). This is because, at each step, the total length of the precursor SMILES sequences remains \(\mathcal{O}(h)\), as the combined number of heavy atoms in the precursors approximates that of the product.  

Thus, for a single molecule, the overall complexity is primarily determined by the sequence length of the synthetic routes, scaling as \(\mathcal{O}(h^4)\) as \(d\) is constant. Additionally, the attention mechanism is applied iteratively for each predicted token until an end token is reached, meaning that the number of tokens scales with the sequence length. Consequently, the computational complexity of DMS is given by \(\mathcal{O}(h^2 \cdot h^4) = \mathcal{O}(h^6)\).

\subsection{FragmentRetro complexity analysis}\label{subsection:complexity_fragmentretro}
The computational complexity of FragmentRetro consists of the following main components:
\begin{enumerate}
    \item \textbf{Fragmentation:} The BRICS (and r-BRICS) algorithm examines all bonds in molecules to generate fragments. The worst case scales \(\mathcal{O}(h^2)\) assuming molecules are fully connected as complete graphs where atoms are nodes and bonds are edges. In practice, there are valency constraints for the number of bonds, for example carbons typically form four bonds, so BRICS effectively scales as \(\mathcal{O}(h)\) for organic molecules.
    
    \item \textbf{Fragment combination:} The number of fragments after BRICS is approximately \(h / \Delta h\), where \(\Delta h\) here is similarly defined as the average reduction in heavy atom count per fragmentation bond. Since the BRICS algorithm only generates acyclic graphs (assuming fragments are nodes and the connecting bonds are edges). The total number of possible fragment combinations, for up to \(N =  h / \Delta h\), has an upper bound of  \(\mathcal{O}(N + (N-1) + \ldots + 1) = \mathcal{O}(N^2) = \mathcal{O}(h^2)\) as \(\Delta h\) is constant. Note that r-BRICS can sometimes generate cyclic graphs, but the number of edges is still practically \(\mathcal{O}(N)\). So the number of possible combinations also scales as \(\mathcal{O}(h^2)\). % The practical cases (effective combinations) due to pruning of redundant combinations and constraints on valid combinations could even speed up the complexity, but it cannot be estimated.

    \item \textbf{Substructure search:} Searching for each fragment combination in the compound stock set involves substructure search, which is typically implemented in cheminformatics libraries like RDKit. The complexity of RDKit's substructure search depends on the matching algorithm. Subgraph isomorphism has exponential complexity in the worst case. However, the use of algorithms like VF2 and Ullman algorithms can significantly bring the practical scenario for fragment molecular graphs to \(\mathcal{O}(|\mathcal{B}| \cdot h^3)\)~\cite{substruct_algo_2004}. This estimate assumes that the fragment sizes are \(\mathcal{O}(h)\), though most fragments are considerably smaller. Besides, the use of SMARTS patterns and fingerprint screening significantly reduces this complexity in practice~\cite{substruct_bench_2012, rdkit_2006}. 

    \item \textbf{Fingerprint screening:} To further optimize substructure search, fingerprint screening is employed as a preprocessing step to rapidly filter out BBs that cannot contain the query substructure. Pattern fingerprints encode the presence or absence of predefined molecular fragments into bit vectors, ensuring no false negatives (where "negative" means no match). This screening step reduces the number of candidate molecules from the compound stock set before the more computationally expensive substructure matching. The number of candidates is reduced by an approximate factor of \(k^h\) (\(k > 1\)). This occurs when each heavy atom feature is hashed into the pattern fingerprint, and the fingerprint dimension is large enough to avoid double hashing. In practical scenarios, $k$ is often around 2 , when each feature eliminates roughly half of the BBs on average. Given these constraints, since \(h\) has a natural upper bound for organic molecules and \(k^h\) grows exponentially, \(h^3 / k^h\) rapidly diminishes to a near-constant value in practice, allowing us to approximate it as a constant \(C_{fp}\). Intuitively, larger fragments require fewer substructure matches since they contain more 1's in their pattern fingerprint bit vectors.
\end{enumerate}

Combining these components, the overall complexity of FragmentRetro is \(\mathcal{O}(h + C_{fp} \cdot |\mathcal{B}| \cdot h^2) = \mathcal{O}(h^2)\), assuming \(|\mathcal{B}|\) and \(k\) are constant. This results in a quadratic complexity, in contrast to the exponential complexity of tree-search methods. Without fingerprint screening, the complexity of FragmentRetro would be \(\mathcal{O}(|\mathcal{B}| \cdot h^3 \cdot h^2) = \mathcal{O}(h^5)\). The efficiency gain is due to fingerprint screening, which significantly reduces the number of BBs that require substructure matching.

\clearpage

\section{Additional case studies}\label{sec:add_case}

\begin{figure}[htbp]
  \centering
  \includegraphics[width=\textwidth]{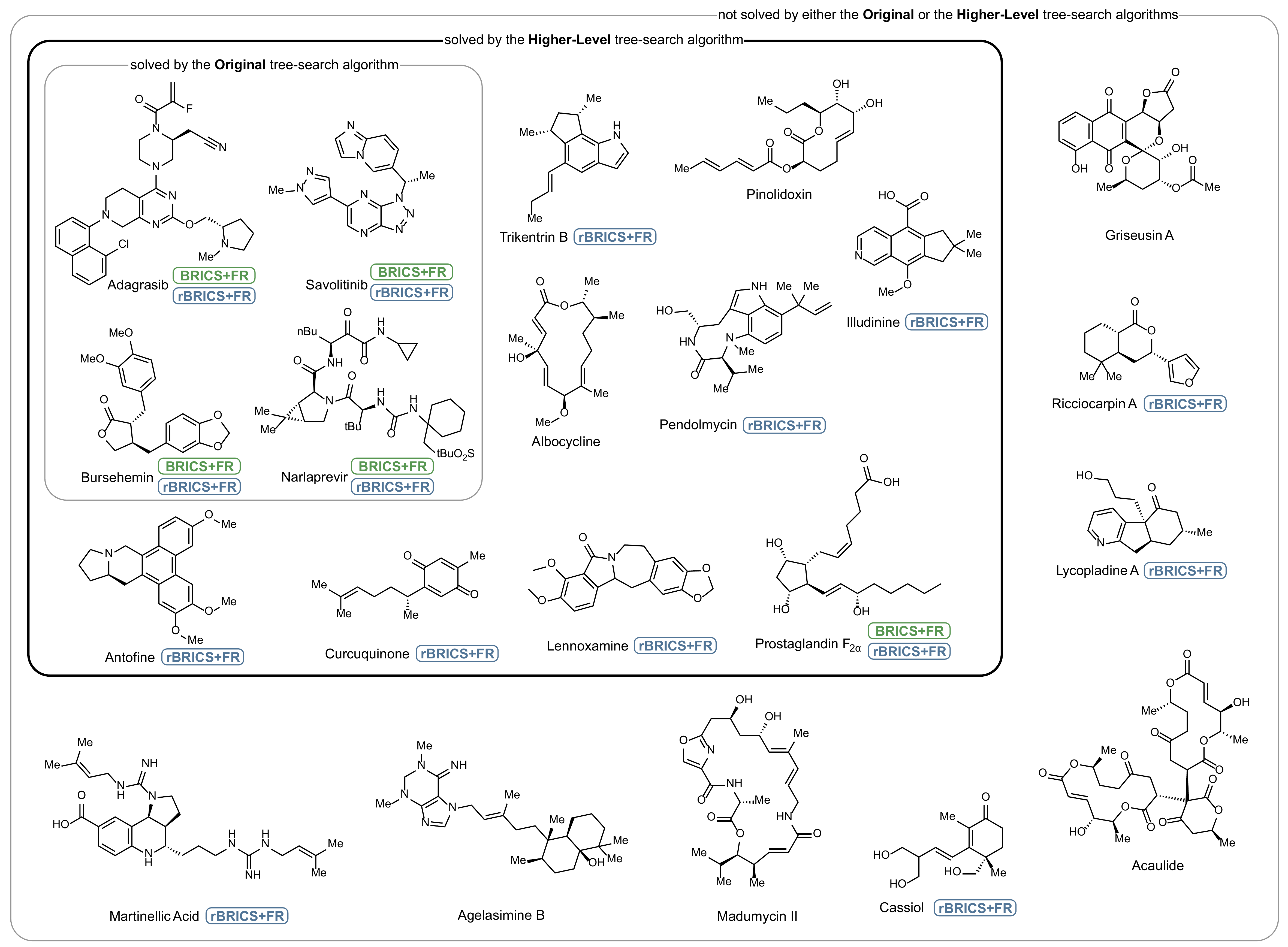}  
  \caption{Additional evaluation of target compounds using FragmentRetro. This figure adopts the format and compound set from Extended Data Fig. 5 of the Higher-Level Retrosynthesis paper~\cite{higherlev_2025} to facilitate direct comparison. The smallest box contains targets solved by the ``Original'' tree-search algorithm (as defined in~\cite{higherlev_2025}); the medium box includes additional compounds solved by the Higher-Level strategy. Compounds outside both boxes were not solved by either method. Compounds successfully solved by BRICS + FragmentRetro and r-BRICS + FragmentRetro are marked with green (BRICS+FR) and blue (r-BRICS+FR) tags, respectively.}
  \label{fig:natural_products_ext}
  \vspace{-1em}
\end{figure}

\end{document}